\documentclass{article}

\usepackage[]{ijcai-2023}

\usepackage{natbib}
\usepackage[utf8]{inputenc} 
\usepackage[T1]{fontenc}    
\usepackage{hyperref}       
\usepackage{url}            
\usepackage{booktabs}       
\usepackage{amsfonts}       
\usepackage{nicefrac}       
\usepackage{microtype}      
\usepackage{xcolor}         
\usepackage{booktabs}
\usepackage{multirow}
\usepackage{mathrsfs}
\usepackage{ amssymb }
\usepackage{mathtools}
\usepackage{comment}
\usepackage{subcaption}
\usepackage{float}

\usepackage{adjustbox}
\title{What Constitutes Good Contrastive Learning \\in Time-Series Forecasting?}

\author{
Chiyu Zhang$^{1,2}$
\and
Qi Yan$^{1,2}$
\and
Lili Meng$^2$\and
Tristan Sylvain$^{2}$
\affiliations
$^1$University of British Columbia\\
$^2$Borealis AI\\
\emails
chiyuzh@mail.ubc.ca~~~qi.yan@ece.ubc.ca~~~\{firstname.lastname\}@borealisai.com
}

\begin{document}

\maketitle

\begin{abstract}
In recent years, the introduction of self-supervised contrastive learning (SSCL) has demonstrated remarkable improvements in representation learning across various domains, including natural language processing and computer vision. By leveraging the inherent benefits of self-supervision, SSCL enables the pre-training of representation models using vast amounts of unlabeled data. Despite these advances, there remains a significant gap in understanding the impact of different SSCL strategies on time series forecasting performance, as well as the specific benefits that SSCL can bring.

This paper aims to address these gaps by conducting a comprehensive analysis of the effectiveness of various training variables, including different SSCL algorithms, learning strategies, model architectures, and their interplay. Additionally, to gain deeper insights into the improvements brought about by SSCL in the context of time-series forecasting, a qualitative analysis of the empirical receptive field is performed. Through our experiments, we demonstrate that the end-to-end training of 
a Transformer model using the Mean Squared Error (MSE) loss and SSCL emerges as the most effective approach in time series forecasting. Notably, the incorporation of the contrastive objective enables the model to prioritize more pertinent information for forecasting, such as scale and periodic relationships. These findings contribute to a better understanding of the benefits of SSCL in time series forecasting and provide valuable insights for future research in this area. Our codes are available at \url{https://github.com/chiyuzhang94/contrastive_learning_time-series_e2e}. 
\end{abstract}

\section{Introduction}\label{sec:intro}
Self-supervised contrastive learning (SSCL) has demonstrated remarkable performance in computer vision (CV)~\citep{he-2020-momentum, chen2020simple, kotar2021contrasting}, natural language processing (NLP)~\citep{gao2021simcse, liu2021fast} and many other domains~\citep{liu2021social, liu2021self}. This family of techniques functions by leveraging unlabeled data to construct positive (similar) samples and negative (non-similar) samples. Recently, SSCL approaches such as TS2Vec~\citep{yue2022ts2vec} and CoST~\citep{woo2022cost} have been applied with success to time series data.

Within the realm of time series forecasting, Transformer-based models~\citep{shabani2022scaleformer, zhou2021informer, wu2021autoformer, zhou2022fedformer} have consistently showcased superior performance compared to Temporal Convolutional Networks (TCNs) when employed as backbones in end-to-end setups. Curiously, despite this clear advantage, mainstream approaches to SSCL for time-series forecasting predominantly utilize TCNs. These TCN-based approaches, while demonstrating robust performance in contrastive learning, raise a fundamental question: Are Transformers unsuitable for effective contrastive learning for time-series forecasting?

Furthermore, SSCL approaches for time series typically follow a two-step process. Initially, a representation is learned on an in-domain dataset using contrastive objectives, and subsequently, a regressor is trained on the acquired representation to address a downstream task with labeled data. Nevertheless, it remains uncertain whether alternative end-to-end approaches possess untapped potential. To gain a comprehensive understanding of the optimal strategies for harnessing SSCL in time-series forecasting and the specific ways in which SSCL enhances its performance, further research is imperative.

Within this study, we offer a thorough investigation into the efficacy of SSCL for time-series forecasting. Our exploration encompasses a diverse array of SSCL algorithms specifically tailored for time series, diverse learning strategies, and backbone architectures. By undertaking such an extensive examination, we aim to establish a holistic comprehension of optimal methods for harnessing SSCL's potential within the realm of time-series forecasting. Our aim in this paper is to provide answers to the following questions:
\begin{enumerate}
    \item Which backbone model is best suited for leveraging SSCL in time-series forecasting models?
    \item What constitutes the optimal learning strategy for effectively employing SSCL for time-series models?
    \item In what ways does the integration of SSCL enhance the performance of time-series forecasting models?
\end{enumerate}
\section{Related Work}

\paragraph{\textbf{Self-Supervised Learning.}}
Supervision data is often limited and expensive to obtain. To alleviate the dependency on labeled data, self-supervised learning has been introduced in various domains. In NLP, the self-supervised pre-training (SSPT) enables models such as BERT~\citep{devlin-2018-bert} to learn effective task-agnostic representations from large amounts of unlabeled data. Similarly, SSPT has also been employed to learn a powerful visual representation model (e.g., ViT~\citep{alexey-2021-vit}) and transfer learned knowledge to a wide range of downstream tasks. Despite the success of SSPT in NLP and CV, its application to time series data~\cite{shi-2021-selfsupervised} remains underexplored due to the potential mismatch between the pre-training and target domains in time series data. Time series data can exhibit large shifts in temporal dynamics, irregular sampling, system factors, etc.~\citep{fawaz2018transfer, ye-2021-implementing}. Recently, a few studies have begun to exploit SSPT in the domain of time series data with contrastive leaning~\cite{yue2022ts2vec,woo2022cost}. 

\paragraph{\textbf{Transformer based model for Time-Series.}}
Transformers~\citep{vaswani2017attention} based models have exhibit remarkable capabilities across a variety of domains and tasks, such as computer vision~\citep{ranftl2021vision} and natural language processing~\citep{bubeck2023sparks}. Recently it transferred the success to the time-series forecasting domain. \citet{tang2021probabilistic} proposes a probabilistic, non-auto-regressive Transformer-based model with the integration of state space models. Informer~\citep{zhou2021informer} lowered the original quadratic complexity in time and memory was lowered to $O(L \log L)$ by enforcing sparsity in the attention mechanism with the ProbSparse attention. While such attention mechanisms operate on a point-wise basis, Autoformer~\citep{wu2021autoformer} used a cross-correlation-based attention mechanism to operate at the level of subsequences. This, along with trend/cycle decomposition resulted in improved performance. FedFormer~\citep{zhou2022fedformer} introduces FFT and wavelet-based blocks to boost performance. ScaleFormer~\citep{shabani2022scaleformer} introduces iterative refinement on different scales to improve performance.
More recently, the PatchTST~\citep{nie2023time} and Crossformer~\citep{zhang2023crossformer} propose to segregate the multivariate time-series data into univariate ones to better capture the multidimensional interaction.

\paragraph{\textbf{SSCL for Time-Series.}}
\citet{yue2022ts2vec} introduces a contrastive learning framework (i.e. TS2Vec) for representation learning on time series data. TS2Vec obtains positive samples via applying timestamp masking and random cropping. TS2Vec also randomly samples two overlapping subseries from the input time series as a positive pair and uses them to encourage consistency of contextual representation. To learn at different semantic levels, TS2Vec trains a TCN model with hierarchical SSCL losses. It applies max-pooling with a kernel size of 2 on the learned representation along the time axis and contrasts representations from the temporal level to instance-level gradually. \citet{woo2022cost} introduces a seasonal-trend contrastive learning framework (i.e., CoST) to learn disentangled seasonal-trend representations. CoST utilizes a TCN model to encode an input time series into an intermediate representation and then disentangles it to trend and seasonal features. CoST applies a contrastive loss for each disentangled representation. In the time domain, they contrast trend representations using the MoCo~\citep{he-2020-momentum} framework to exploit a large number of negative samples. For the frequency domain, CoST applies a contrastive loss on the amplitude and phase representations of each frequency, respectively. After contrastive pre-training, both TS2Vec and CoST keep the trained model frozen, use it as a feature extractor to encode samples of a downstream task into latent representations, and use the learned representations to train a regressor (e.g., SVM) for time series forecasting. 
However, the identification of appropriate augmentation methods and negative pairs for time-series forecasting remains challenging. 
A contemporaneous work by \cite{zheng-2023-simts} introduces SimTS framework to learn time series representations by predicting its future from history. This framework incorporates a siamese network, a multiscale encoder, and a predictor network. 
SimTS does not require negative pairs and it makes no assumptions about the characteristics of individual time series.
\section{What constitutes good contrastive learning for time series forecasting?}
Different from the pre-training in CV and NLP, the pre-training of time series usually uses the same dataset as the downstream task but without exploiting labels. Given a time-series dataset $\mathcal{D}=\{x_1, x_2, \dots, x_N\}$ with $N$ instances, we train an encoder $\Phi(\cdot)$ to map each $x_i$ in its representation $r_i$ that can capture effective information for time-series forecasting, where each $x_i \in \mathcal{R}^{t \times m}$, $r_i \in \mathcal{R}^{t \times d}$, $t$ is the timestamp size, $m$ is the dimension of input signals, and $d$ is the hidden size of representations.  

\subsection{Which backbone model is the best?} 
We investigate the following three backbone architectures: \textbf{(1) Long short-term memory network (LSTM)}~\citep{hochreiter1997long} is a variant of RNN, which processes input sequence recursively and regularizes information flow with ``gates" (i.e., forget and cell gates). \textbf{(2) Temporal Convolutional Network (TCN)}~\citep{bai-2018-empirical} is a convolutional neural network for processing sequential data, which includes casual convolutions and dilated convolutions. Dilated convolutions enable TCNs to capture long-range dependencies in the input data while also maintaining a relatively small number of parameters. \textbf{(3) Transformer}~\citep{vaswani2017attention} is a fully attention-based architecture. A Transformer consists of an input embedding layer and one or more Transformers layers. Each Transformers layer has a multi-head self-attention sublayer and a feed-forward sublayer. Self-attention enables Transformers to capture both local and global dependencies in the sequential data. In our experiments, we utilize Informer~\cite{zhou-2021-informer} architecture that replaces vanilla multi-head attention with a ProbSparse self-attention to improve the model's efficiency. 

\subsection{Which learning strategy is the best?}

\paragraph{End-to-End Training.} To train a model for time series forecasting, an MLP is added on top of the encoder $\Phi(\cdot)$, which takes the last $T$ hidden representations $\{r_{N-T}, r_{N-T+1}, \dots, r_{N} \}$ from $\Phi(\cdot)$ to predict the future $T$ observations $\{x_{N+1}, x_{N+2}, \dots , x_{N+T}\}$. The MLP head and encoder are trained on a supervised dataset with MSE loss (Eq.~\ref{eq:mse}). We incorporate SSCL into an end-to-end training setup by using SSCL as an auxiliary objective. SSCL takes all representations $\{r_{1}, r_{2}, \dots, r_{N}\}$ to calculate the corresponding contrastive loss. We introduce a scale $\lambda$ to balance the MSE loss and contrastive loss. 

\begin{equation}
    \mathcal{L}_{i}^{MSE} = (x_i - \hat{x_i})^2 \label{eq:mse}
\end{equation}

\paragraph{Two-Step Training.} Previous work pre-trains an encoder $\Phi(\cdot)$ with SSCL first and then uses the pre-trained encoder as a feature extractor (i.e., keeping $\Phi(\cdot)$ frozen) to map input time series into latent representations. Different from the pre-training in NLP and CV, a time series encoder is usually pre-trained on the same dataset as the downstream task dataset due to the severe domain shifting issue in time series data. \textbf{(1) Regressor}. TS2Vec and CoST utilize the learned representations to train an external train linear regression model with L2 norm penalty to predict future $\hat{x} = \{\hat{x}_{N+1}, \hat{x}_{N+2}, \dots , \hat{x}_{N+T}\}$. \textbf{(2) MLP}. We train an MLP with MSE loss to predict $\hat{x}$ while using $\Phi(\cdot)$ as a frozen model. \textbf{(3) Fine-tuning.} While pre-training and fine-tuning learning framework has achieved impressive improvement on NLP and CV tasks, existing studies have not investigated its utility in time series forecasting. Thus, we add a new MLP head on top of the pre-trained encoder $\Phi(\cdot)$ and fine-tune them with MSE loss end-to-end. 

\subsection{Which loss function is the best?}
We investigate two main SSCL algorithms and their variants. For all the algorithms, they utilize InfoNCE contrastive loss~\citep{chen2017sampling} but construct negative samples differently. Eq.~\ref{eq:sscl} shows InfoNCE loss, where $sim(\cdot)$ is cosine similarity $\frac{h_i^\top h_j}{\|h_i\|\cdot\|h_j\|}$, $h_i$, $h_{p(i)}$, and $h_a$ are representations of an anchor, positive sample, and negative sample, respectively, and $Neg$ is a set of negative samples. \textbf{(1) Hierarchical Contrastive Loss (HCL)}~\citep{yue2022ts2vec} aggregates representations of timestamps by using a max pooling with a kernel size of 2 and applies contrastive loss along the time axis, recursively. At the end of aggregation, i.e., each time series is represented in a vector, the model can also learn instance-level representations. \textbf{(2) Momentum Contrast (MoCo1)}~\citep{he-2020-momentum} utilizes a momentum encoder to produce a positive sample and a dynamic memory bank to provide a large size of negative samples.
MoCo provides an instance-level contrastive loss, which randomly selects a time stamp to represent the whole sequence.
\textbf{(3) MoCo2}~\citep{chen-2020-improved} involves three components into MoCo1 to enhance SSCL, i.e., (i) replacing a feed-forward projection layer with a two-layered MLP with ReLU; (ii) more data augmentation techniques~\footnote{We use masking, jitter, scale, and shift.}; and (iii) a cosine learning rate scheduler. \textbf{(4) HCL+MoCo2.} We then combine the losses from HCL and MoCo with equal weight.   

\begin{equation}
	\small 
    \mathcal{L}_{i}^{SSCL} = - \log \frac{e^{sim(h_i, h_{p(i)})/\tau}}{e^{sim(h_i, h_{p(i)})/\tau} +\sum_{a\in Neg} e^{sim(h_i, h_a)/ \tau}}\label{eq:sscl}
\end{equation}

\section{Experiments}
\subsection{Datasets}
We conduct experiments on three real-world public datasets: the electricity transformer temperature (ETT) dataset~\citep{zhou-2021-informer}, which includes an hourly-level dataset (ETTh) and a 15-minute-level dataset (ETTm), and the electricity dataset (ECL)\footnote{\url{https://archive.ics.uci.edu/ml/datasets/ElectricityLoadDiagrams20112014}} which contains the electricity consumption of 312 clients.
The ETT datasets consist of data from six power load features and measurements of oil temperature, while the ECL dataset was converted into hourly-level measurements, following previous work~\citep{yue2022ts2vec}. 
We conduct our experiments for multivariate forecasting and use MSE as the main evaluation metric, while also reporting the MAE results.
To ensure an overall evaluation, we average the MSE across all prediction lengths and the three datasets to represent the overall performance of each model.

\subsection{Implementation details}
\begin{figure}
  \centering
  \begin{subfigure}{0.45\textwidth}
    \centering
    \begin{adjustbox}{width=\textwidth}
    \includegraphics{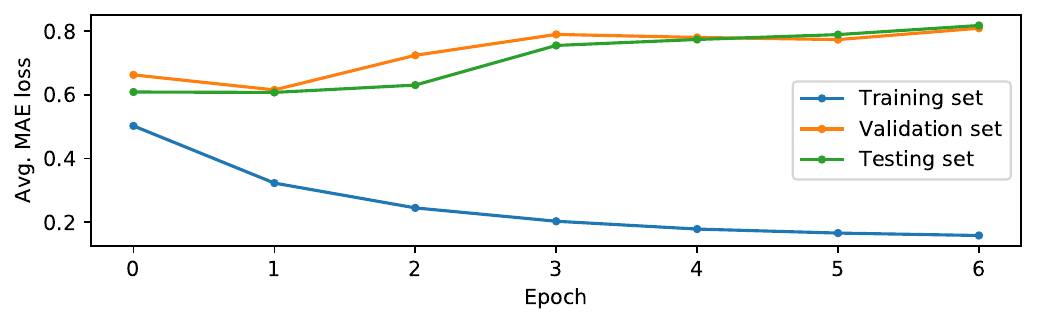}
    \end{adjustbox}
    \vspace{-1.5em}
    \caption{Informer w/ early stopping.}
    \vspace{-0.2em}
  \end{subfigure}
    \begin{subfigure}{0.45\textwidth}
    \centering
    \begin{adjustbox}{width=\textwidth}
    \includegraphics{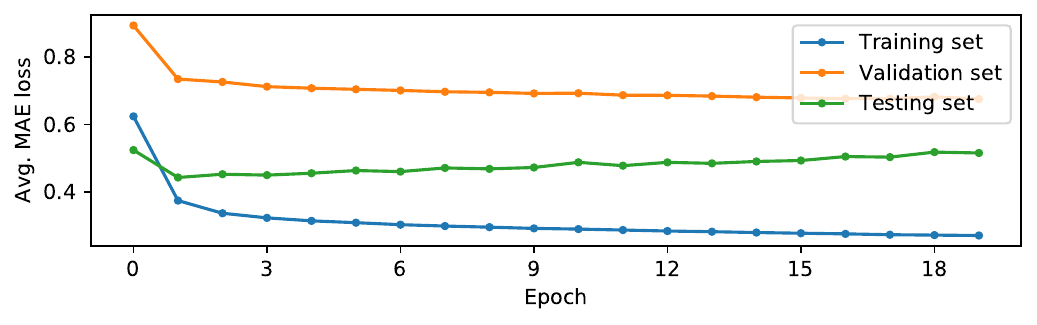}
    \end{adjustbox}
    \vspace{-1.5em}
    \caption{Informer w/o early stopping.}
    \vspace{-0.2em}
  \end{subfigure}
  \hfill
  \begin{subfigure}{0.45\textwidth}
    \centering
    \begin{adjustbox}{width=\textwidth}
      \includegraphics{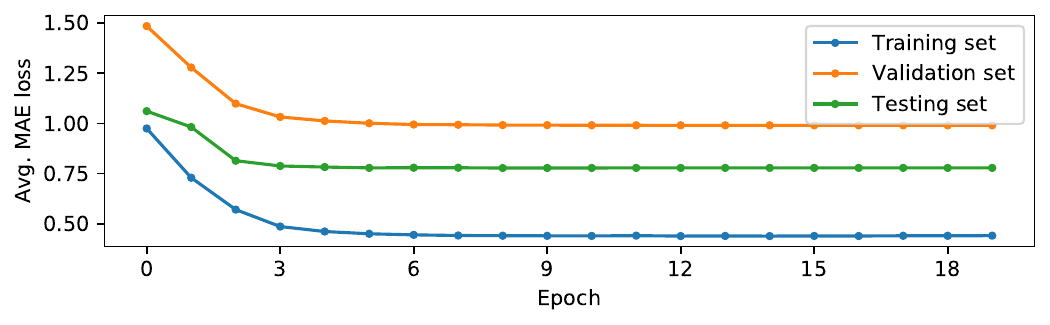}
    \end{adjustbox}
    \vspace{-1.5em}
    \caption{TCN w/ early stopping.}
    \vspace{-0.2em}
  \end{subfigure}
  \vspace{-0.2em}
  \caption{Training curves of Informer and TCN backbones on the ETTh1 dataset (prediction horizon $24$).
  }
  \vspace{-1.0em}
  \label{fig:training_curve}
\end{figure}

We implement our models with PyTorch. The LSTM-based encoder includes five uni-directional LSTM layers with a hidden dimension of 320 in each layer. This architecture has a total of 660K learnable parameters. The TCN-based encoder consists of 10 dilated convolution layers with a hidden dimension of 320 and a kernel size of 3 per layer. This architecture has a total of 637K learnable parameters. The Transformer-based encoder contains 5 Informer layers with a hidden size of 128 and 8 attention heads. This architecture has a total of 655K learnable parameters. In all three models, we utilize a linear layer as the first embedding layer to map the input features into hidden dimensions. We train all three models with a peak learning rate of 0.001, and use a cosine learning rate scheduler for MoCo2-framework models. For end-to-end training, each model trains for 30 epochs with early stopping 
based on the Dev performance to prevent overfitting. For two-step learning, we train an encoder on the training set with SSCL for 600 training iterations. When fine-tuning a pre-trained encoder, we use the same training hyper-parameters as in end-to-end training.

\begin{table*}
\centering
\setlength{\tabcolsep}{4pt}
\tiny
\begin{tabular}{@{}lrcccccccccc|cccccc@{}}
\toprule
\textbf{} & & \multicolumn{10}{c}{\textbf{End-to-End Learning}}   & \multicolumn{6}{c}{\textbf{Two-Step Learning}}  \\ \cmidrule(l){3-12} \cmidrule(l){13-18}
\textbf{} & & \multicolumn{2}{c}{\textbf{MSE}} & \multicolumn{2}{c}{\textbf{+HCL}} & \multicolumn{2}{c}{\textbf{+MoCo}} & \multicolumn{2}{c}{\textbf{+MoCo2}} & \multicolumn{2}{c|}{\textbf{+MoCo2+HCL}} & \multicolumn{2}{c}{\textbf{EC+MLP}} & \multicolumn{2}{c}{\textbf{EC+SVM}} & \multicolumn{2}{c}{\textbf{FT}} \\ \cmidrule(l){3-4} \cmidrule(l){5-6} \cmidrule(l){7-8} \cmidrule(l){9-10} \cmidrule(l){11-12} \cmidrule(l){13-14} \cmidrule(l){15-16} \cmidrule(l){17-18}
\textbf{} & & \textbf{MSE} & \textbf{MAE} & \textbf{MSE} & \textbf{MAE} & \textbf{MSE} & \textbf{MAE} & \textbf{MSE} & \textbf{MAE} & \textbf{MSE} & \textbf{MAE} & \textbf{MSE} & \textbf{MAE} & \textbf{MSE} & \textbf{MAE} & \textbf{MSE} & \textbf{MAE} \\ \midrule
\multirow{5}{*}{\textbf{ECL}} & 24 & 0.305 & 0.372 & 0.317 & 0.383 & 0.305 & 0.376 & 0.304 & 0.371 & 0.302 & 0.372 & 1.011 & 0.830 & 1.020 & 0.820 & 0.301 & 0.372 \\
 & 48 & 0.355 & 0.409 & 0.325 & 0.391 & 0.374 & 0.421 & 0.421 & 0.428 & 0.383 & 0.416 & 1.010 & 0.830 & 1.019 & 0.820 & 0.388 & 0.424 \\
 & 168  & 0.324 & 0.390 & 0.354 & 0.419 & 0.316 & 0.388 & 0.324 & 0.395 & 0.331 & 0.398 & 1.010 & 0.830 & 1.019 & 0.820 & 0.325 & 0.391 \\
 & 336  & 0.367 & 0.419 & 0.345 & 0.410 & 0.355 & 0.416 & 0.370 & 0.420 & 0.342 & 0.400 & 1.006 & 0.828 & 1.015 & 0.820 & 0.376 & 0.422 \\
 & 720  & 0.401 & 0.425 & 0.391 & 0.441 & 0.347 & 0.404 & 0.383 & 0.421 & 0.396 & 0.433 & 0.999 & 0.826 & 1.009 & 0.819 & 0.398 & 0.431 \\\hline
\multirow{5}{*}{\textbf{ETTh1}} & 24 & 0.912 & 0.757 & 0.934 & 0.756 & 1.000 & 0.790 & 0.950 & 0.792 & 0.927 & 0.756 & 1.109 & 0.795 & 1.107 & 0.797 & 0.991 & 0.800 \\
 & 48 & 0.979 & 0.788 & 1.082 & 0.804 & 1.010 & 0.804 & 1.030 & 0.810 & 1.029 & 0.804 & 1.110 & 0.797 & 1.106 & 0.798 & 1.063 & 0.836 \\
 & 168  & 1.151 & 0.845 & 1.178 & 0.847 & 1.042 & 0.811 & 0.991 & 0.781 & 1.022 & 0.803 & 1.108 & 0.801 & 1.104 & 0.799 & 1.060 & 0.823 \\
 & 336  & 1.123 & 0.844 & 1.362 & 0.920 & 1.051 & 0.820 & 1.038 & 0.819 & 1.064 & 0.822 & 1.089 & 0.801 & 1.094 & 0.799 & 1.053 & 0.804 \\
 & 720  & 1.069 & 0.834 & 1.148 & 0.854 & 1.140 & 0.855 & 1.203 & 0.866 & 1.135 & 0.844 & 1.068 & 0.810 & 1.072 & 0.797 & 1.183 & 0.865 \\\hline
\multirow{5}{*}{\textbf{ETTm1}} & 24 & 1.064 & 0.784 & 1.076 & 0.810 & 1.075 & 0.796 & 0.991 & 0.743 & 1.070 & 0.798 & 1.107 & 0.796 & 1.089 & 0.787 & 1.043 & 0.720 \\
 & 48 & 0.980 & 0.756 & 0.986 & 0.777 & 0.957 & 0.751 & 0.975 & 0.779 & 0.866 & 0.676 & 1.107 & 0.793 & 1.095 & 0.789 & 1.014 & 0.734 \\
 & 96 & 0.865 & 0.724 & 1.057 & 0.793 & 0.841 & 0.715 & 0.893 & 0.741 & 0.856 & 0.723 & 1.108 & 0.794 & 1.091 & 0.788 & 0.925 & 0.730 \\
 & 288  & 1.091 & 0.818 & 1.048 & 0.800 & 1.165 & 0.847 & 1.017 & 0.796 & 1.045 & 0.800 & 1.104 & 0.796 & 1.093 & 0.790 & 1.466 & 0.963 \\
 & 672  & 1.039 & 0.790 & 1.221 & 0.850 & 1.144 & 0.831 & 1.104 & 0.822 & 1.097 & 0.811 & 1.102 & 0.797 & 1.095 & 0.793 & 1.215 & 0.854 \\\hline
\textbf{Average} & & 0.802 & 0.664 & 0.855 & 0.684 & 0.808 & 0.668 & 0.800 & 0.666 & \textbf{0.791} & \textbf{0.657} & 1.070 & 0.808 & 1.068 & 0.802 & \textbf{0.853} & \textbf{0.678} \\ \bottomrule
\end{tabular}
\vspace{-1.0em}
\caption{Multivariate time series forecasting results of LSTM-based models.}\label{tab:lstm}
\vspace{-0.5em}
\end{table*}
\begin{table*}[]
\centering
\setlength{\tabcolsep}{4pt}
\tiny
\begin{tabular}{@{}lrcccccccccc|cccccc@{}}
\toprule
\textbf{} & & \multicolumn{10}{c}{\textbf{End-to-End Learning}}   & \multicolumn{6}{c}{\textbf{Two-Step Learning}}  \\ \cmidrule(l){3-12} \cmidrule(l){13-18}
\textbf{} & & \multicolumn{2}{c}{\textbf{MSE}} & \multicolumn{2}{c}{\textbf{+HCL}} & \multicolumn{2}{c}{\textbf{+MoCo}} & \multicolumn{2}{c}{\textbf{+MoCo2}} & \multicolumn{2}{c|}{\textbf{+MoCo2+HCL}} & \multicolumn{2}{c}{\textbf{EC+MLP}} & \multicolumn{2}{c}{\textbf{EC+SVM}} & \multicolumn{2}{c}{\textbf{FT}} \\ \cmidrule(l){3-4} \cmidrule(l){5-6} \cmidrule(l){7-8} \cmidrule(l){9-10} \cmidrule(l){11-12} \cmidrule(l){13-14} \cmidrule(l){15-16} \cmidrule(l){17-18}
\textbf{} & & \textbf{MSE} & \textbf{MAE} & \textbf{MSE} & \textbf{MAE} & \textbf{MSE} & \textbf{MAE} & \textbf{MSE} & \textbf{MAE} & \textbf{MSE} & \textbf{MAE} & \textbf{MSE} & \textbf{MAE} & \textbf{MSE} & \textbf{MAE} & \textbf{MSE} & \textbf{MAE} \\ \midrule
\multirow{5}{*}{\textbf{ECL}} & 24 & 0.263 & 0.353 & 0.288 & 0.364 & 0.262 & 0.351 & 0.268 & 0.354 & 0.259 & 0.350 & 0.305 & 0.388 & 0.526 & 0.537 & 0.270 & 0.356 \\
 & 48 & 0.283 & 0.366 & 0.318 & 0.392 & 0.308 & 0.382 & 0.306 & 0.383 & 0.291 & 0.371 & 0.335 & 0.407 & 0.538 & 0.543 & 0.304 & 0.379 \\
 & 168 & 0.298 & 0.380 & 0.335 & 0.411 & 0.307 & 0.387 & 0.318 & 0.398 & 0.318 & 0.397 & 0.298 & 0.385 & 0.545 & 0.547 & 0.305 & 0.383 \\
 & 336 & 0.287 & 0.367 & 0.360 & 0.425 & 0.292 & 0.377 & 0.292 & 0.374 & 0.296 & 0.376 & 0.330 & 0.409 & 0.544 & 0.548 & 0.297 & 0.375 \\
 & 720 & 0.308 & 0.380 & 0.362 & 0.424 & 0.360 & 0.417 & 0.352 & 0.413 & 0.361 & 0.420 & 0.393 & 0.448 & 0.547 & 0.552 & 0.354 & 0.415 \\ \hline
\multirow{5}{*}{\textbf{ETTh1}} & 24 & 0.641 & 0.600 & 0.884 & 0.724 & 0.524 & 0.507 & 0.515 & 0.508 & 0.506 & 0.499 & 0.562 & 0.511 & 0.890 & 0.686 & 0.536 & 0.526 \\
 & 48 & 0.631 & 0.614 & 0.889 & 0.717 & 0.528 & 0.535 & 0.526 & 0.526 & 0.511 & 0.516 & 0.512 & 0.500 & 0.935 & 0.712 & 0.529 & 0.530 \\
 & 168 & 0.821 & 0.712 & 0.999 & 0.768 & 0.818 & 0.702 & 0.798 & 0.677 & 0.770 & 0.680 & 0.718 & 0.624 & 1.023 & 0.766 & 0.823 & 0.702 \\
 & 336 & 1.049 & 0.792 & 1.110 & 0.820 & 0.932 & 0.752 & 0.918 & 0.742 & 0.962 & 0.761 & 0.893 & 0.737 & 1.079 & 0.802 & 0.930 & 0.725 \\
 & 720 & 1.321 & 0.889 & 1.129 & 0.836 & 0.977 & 0.776 & 0.979 & 0.771 & 0.938 & 0.758 & 1.012 & 0.813 & 1.090 & 0.817 & 0.896 & 0.740 \\\hline
\multirow{5}{*}{\textbf{ETTm1}} & 24 & 0.547 & 0.525 & 0.941 & 0.756 & 0.520 & 0.530 & 0.519 & 0.536 & 0.524 & 0.540 & 0.957 & 0.713 & 0.611 & 0.540 & 0.610 & 0.558 \\
 & 48 & 0.700 & 0.576 & 0.875 & 0.711 & 0.659 & 0.590 & 0.636 & 0.561 & 0.653 & 0.565 & 1.055 & 0.764 & 0.810 & 0.646 & 0.657 & 0.588 \\
 & 96 & 0.624 & 0.582 & 0.983 & 0.775 & 0.503 & 0.501 & 0.501 & 0.505 & 0.510 & 0.508 & 0.518 & 0.497 & 0.874 & 0.680 & 0.528 & 0.537 \\
 & 288 & 1.013 & 0.790 & 1.145 & 0.850 & 0.931 & 0.736 & 0.711 & 0.638 & 0.748 & 0.652 & 0.628 & 0.580 & 0.933 & 0.718 & 0.891 & 0.737 \\
 & 672 & 1.151 & 0.820 & 1.025 & 0.794 & 0.839 & 0.713 & 0.821 & 0.709 & 0.841 & 0.707 & 0.715 & 0.629 & 0.995 & 0.756 & 0.915 & 0.737 \\\hline
\textbf{Average} & & 0.663 & 0.583 & 0.776 & 0.651 & 0.584 & 0.550 & \textbf{0.564} & \textbf{0.540} & 0.566 & \textbf{0.540} & 0.615 & 0.560 & 0.796 & 0.657 & \textbf{0.590} & \textbf{0.552} \\ \bottomrule
\end{tabular}
\vspace{-1.0em}
\caption{Multivariate time series forecasting results of TCN-based models.}\label{tab:tcn}
\vspace{-0.5em}
\end{table*}
\begin{table*}[]
	\setlength{\tabcolsep}{4pt}
	\tiny
	\begin{tabular}{@{}lrcccccccccc|cccccc@{}}
		\toprule
		\textbf{} & & \multicolumn{10}{c}{\textbf{End-to-End Learning}}   & \multicolumn{6}{c}{\textbf{Two-Step Learning}}  \\ \cmidrule(l){3-12} \cmidrule(l){13-18}
		\textbf{} & & \multicolumn{2}{c}{\textbf{MSE}} & \multicolumn{2}{c}{\textbf{+HCL}} & \multicolumn{2}{c}{\textbf{+MoCo}} & \multicolumn{2}{c}{\textbf{+MoCo2}} & \multicolumn{2}{c|}{\textbf{+MoCo2+HCL}} & \multicolumn{2}{c}{\textbf{EC+MLP}} & \multicolumn{2}{c}{\textbf{EC+SVM}} & \multicolumn{2}{c}{\textbf{FT}} \\ \cmidrule(l){3-4} \cmidrule(l){5-6} \cmidrule(l){7-8} \cmidrule(l){9-10} \cmidrule(l){11-12} \cmidrule(l){13-14} \cmidrule(l){15-16} \cmidrule(l){17-18}
		\textbf{} & & \textbf{MSE} & \textbf{MAE} & \textbf{MSE} & \textbf{MAE} & \textbf{MSE} & \textbf{MAE} & \textbf{MSE} & \textbf{MAE} & \textbf{MSE} & \textbf{MAE} & \textbf{MSE} & \textbf{MAE} & \textbf{MSE} & \textbf{MAE} & \textbf{MSE} & \textbf{MAE} \\ \midrule
		\multirow{5}{*}{\textbf{ECL}} & 24 & 0.258 & 0.355 & 0.326 & 0.399 & 0.269 & 0.361 & 0.255 & 0.353 & 0.253 & 0.353 & 0.275 & 0.367 & 0.460 & 0.501 & 0.259 & 0.354 \\
		& 48 & 0.286 & 0.369 & 0.335 & 0.396 & 0.284 & 0.371 & 0.285 & 0.371 & 0.284 & 0.370 & 0.311 & 0.389 & 0.474 & 0.508 & 0.291 & 0.372 \\
		& 168 & 0.264 & 0.358 & 0.316 & 0.397 & 0.269 & 0.365 & 0.273 & 0.368 & 0.270 & 0.361 & 0.283 & 0.380 & 0.487 & 0.515 & 0.267 & 0.362 \\
		& 336 & 0.291 & 0.374 & 0.321 & 0.391 & 0.286 & 0.377 & 0.289 & 0.375 & 0.290 & 0.373 & 0.299 & 0.388 & 0.494 & 0.520 & 0.284 & 0.377 \\
		& 720 & 0.352 & 0.411 & 0.386 & 0.421 & 0.360 & 0.413 & 0.347 & 0.408 & 0.360 & 0.414 & 0.408 & 0.447 & 0.504 & 0.529 & 0.365 & 0.419 \\ \hline
		\multirow{5}{*}{\textbf{ETTh1}} & 24 & 0.530 & 0.532 & 0.493 & 0.501 & 0.524 & 0.507 & 0.515 & 0.535 & 0.541 & 0.553 & 0.477 & 0.496 & 0.659 & 0.556 & 0.489 & 0.499 \\
		& 48 & 0.663 & 0.624 & 0.729 & 0.631 & 0.528 & 0.535 & 0.535 & 0.533 & 0.537 & 0.540 & 0.519 & 0.522 & 0.694 & 0.582 & 0.518 & 0.524 \\
		& 168 & 0.806 & 0.679 & 1.136 & 0.832 & 0.818 & 0.702 & 0.750 & 0.655 & 0.750 & 0.644 & 0.710 & 0.622 & 0.847 & 0.674 & 0.733 & 0.631 \\
		& 336 & 1.113 & 0.823 & 1.313 & 0.929 & 0.932 & 0.752 & 0.840 & 0.706 & 0.870 & 0.708 & 0.841 & 0.717 & 0.956 & 0.737 & 0.897 & 0.739 \\
		& 720 & 1.015 & 0.808 & 1.046 & 0.801 & 0.977 & 0.776 & 0.979 & 0.783 & 0.982 & 0.793 & 1.033 & 0.835 & 1.114 & 0.822 & 1.017 & 0.830 \\ \hline
		\multirow{5}{*}{\textbf{ETTm1}} & 24 & 0.890 & 0.692 & 1.045 & 0.734 & 0.520 & 0.530 & 0.523 & 0.526 & 0.545 & 0.524 & 0.528 & 0.540 & 0.489 & 0.462 & 0.522 & 0.529 \\
		& 48 & 0.701 & 0.574 & 0.785 & 0.611 & 0.659 & 0.590 & 0.680 & 0.602 & 0.681 & 0.570 & 0.591 & 0.530 & 0.571 & 0.515 & 0.582 & 0.525 \\
		& 96 & 0.735 & 0.613 & 0.645 & 0.604 & 0.503 & 0.501 & 0.489 & 0.535 & 0.522 & 0.532 & 0.517 & 0.515 & 0.611 & 0.539 & 0.539 & 0.561 \\
		& 288 & 0.881 & 0.677 & 1.000 & 0.769 & 0.931 & 0.736 & 0.605 & 0.568 & 0.606 & 0.560 & 0.701 & 0.825 & 0.681 & 0.587 & 0.715 & 0.832 \\
		& 672 & 0.918 & 0.712 & 0.954 & 0.754 & 0.839 & 0.713 & 0.734 & 0.645 & 0.759 & 0.639 & 0.825 & 0.712 & 0.783 & 0.649 & 0.810 & 0.716 \\ \hline
		\textbf{Average} & & {0.647} & {0.573} & 0.722 & 0.612 & {0.580} & {0.549} & \textbf{0.540} & 0.531 & 0.550 & \textbf{0.529} & 0.554 & 0.552 & {0.655} & {0.525} & \textbf{0.552} & \textbf{0.551} \\ \bottomrule 
	\end{tabular}
	\vspace{-1.0em}
	\caption{Multivariate time series forecasting results of Informer-based models.}\label{tab:informer}
	\vspace{-0.5em}
\end{table*}
\subsection{Quantitative Analysis}
In this section, we aim to answer the questions mentioned in Section~\ref{sec:intro}. Our objective is to carry out a well-regulated study to answer each question. 

\paragraph{\textbf{Which backbone model works best for SSCL?}} 
To determine the most effective backbone architecture for exploiting SSCL in time series forecasting, we conduct experiments with three different encoder architectures: LSTM, TCN, and Informer For each architecture, we train the model end-to-end with both MSE and MoCo2 losses, using SSCL as an auxiliary objective.
Our results indicate that the Transformer-based model outperforms the other two LSTM- and TCN-based architectures, achieving a MSE of $0.540$. The Transformer-based model achieves a reduction of $0.260$ and $0.024$ MSE compared to LSTM-based and TCN-based models, respectively. 

We observe that the performance of the Transformer backbone is more sensitive to the optimization hyperparameters used during training.
The early stopping strategy, which starts with a relatively large initial learning rate, does not yield satisfactory results for the Informer backbone, as shown in Fig.~\ref{fig:training_curve}.
In such cases, the training process may not converge effectively. However, it is worth noting that this strategy tends to be beneficial for other backbones. Properly training the network and realizing its full capacity heavily relies on addressing this sensitivity to the learning rate.

\paragraph{\textbf{Which learning strategy or SSCL algorithm works best for SSCL?}}
To determine the most effective learning strategy for SSCL, we compare different approaches using Informer-based models within the MoCo2 framework.
We test the end-to-end learning strategy and a two-step learning strategy. Our results, shown in Table~\ref{tab:tcn}, indicate that end-to-end learning with SSCL as an auxiliary objective produces better performance, with an average MSE of $0.540$. This model also outperforms one trained solely with MSE loss. In end-to-end learning, we find that MoCo2 proves to be the most effective SSCL algorithm. In our comparison of the three approaches within the two-step learning strategy, we observe that fine-tuning produces the best results (MSE=$0.552$), outperforming the other two encode-frozen approaches.

\paragraph{How do backbones affect learning objectives or learning strategies?}
We now evaluate the effectiveness of SSCL in improving the performance of time series forecasting models with different backbone architectures. First, we compare different end-to-end training approaches, which include training with either solely MSE loss or joint losses (i.e., MSE and SSCL losses).
Our results show that the LSTM-based model obtains a marginal improvement with MoCo2 ($0.002$ MSE decreasing) and MoCo2+HCL ($0.011$ MSE decreasing). Furthermore, our investigation reveals that the TCN-based models yield substantial MSE reductions when integrated with MoCo2 and MoCo2+HCL. Specifically, the TCN-based models achieve reductions of $0.099$ and $0.097$ in MSE with MoCo2 and MoCo2+HCL, respectively. 
Among the various backbone models, the most effective approach involves training the Informer model with MSE and MoCo2 jointly. The Informer framework demonstrates superior performance with the best MSE metric of $0.540$ ($0.024$ lower than TCN) and the best MAE metric of $0.529$ ($0.011$ lower than TCN), respectively.
These findings highlight the importance of considering the specific backbone architecture when implementing SSCL. While some models may benefit from SSCL, others might not show any noticeable improvement in performance.

Next, we compare various two-step learning approaches using different backbone architectures.
Comparing results in Tables~\ref{tab:lstm}, \ref{tab:tcn}, and~\ref{tab:informer}, we can see that the Transformer encoder outperforms the other two backbone models when models are trained with pre-training and fine-tuning framework. The LSTM-, TCN-, and Transformer-based models attain test MSE values of $0.853$, $0.590$, and $0.552$, respectively. 
Across all backbones, the fine-tuning approach is found to be more effective than using a pre-trained model as a frozen encoder. 
Also, the LSTM-based model exhibits poor performance when used as a frozen feature encoder after pre-training (MSE$=1.070$). 
The SVM regressor consistently yields inferior results compared to the MLP prediction head.


Furthermore, our evaluation of different learning strategies reveals that the best performing end-to-end learning model consistently outperforms the best two-step learning models, irrespective of the backbone used.
We validate that adopting end-to-end learning with SSCL as an auxiliary objective on Transformer model is the preferred approach.
This method not only delivers superior performance but also proves more efficient and straightforward to implement.

\begin{figure}[H]
\centering
\begin{subfigure}[]{.49\textwidth}
  \centering
  \includegraphics[width=\linewidth]{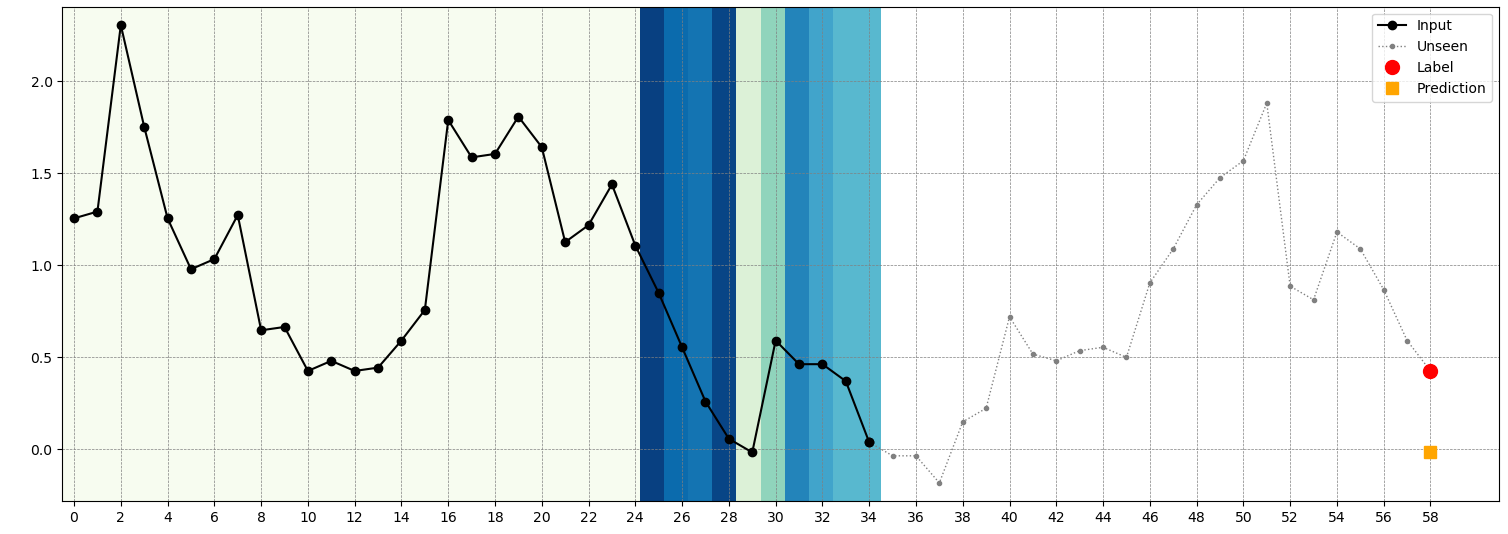}
  \caption{TCN trained with MSE 
  }
  \label{fig:suba}
\end{subfigure}%

\begin{subfigure}[]{.49\textwidth}
  \centering
  \includegraphics[width=\linewidth]{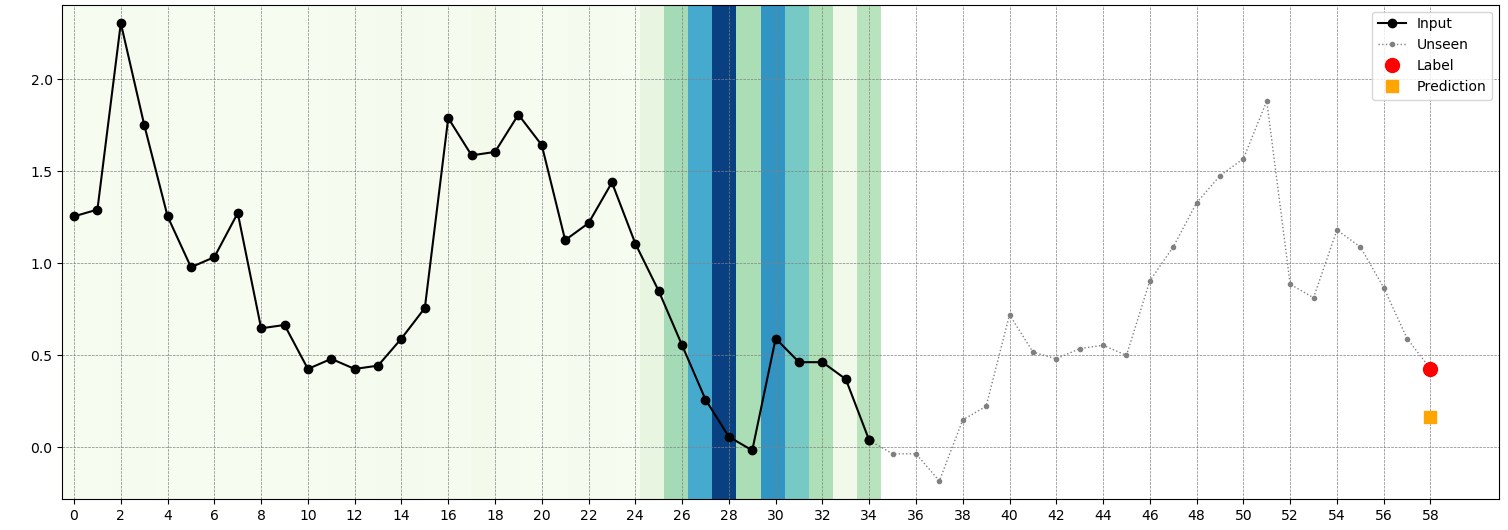}
  \caption{TCN trained with MSE and MoCo2 
  	}\label{fig:subb}
\end{subfigure}%

\begin{subfigure}[]{.49\textwidth}
  \centering
  \includegraphics[width=\linewidth]{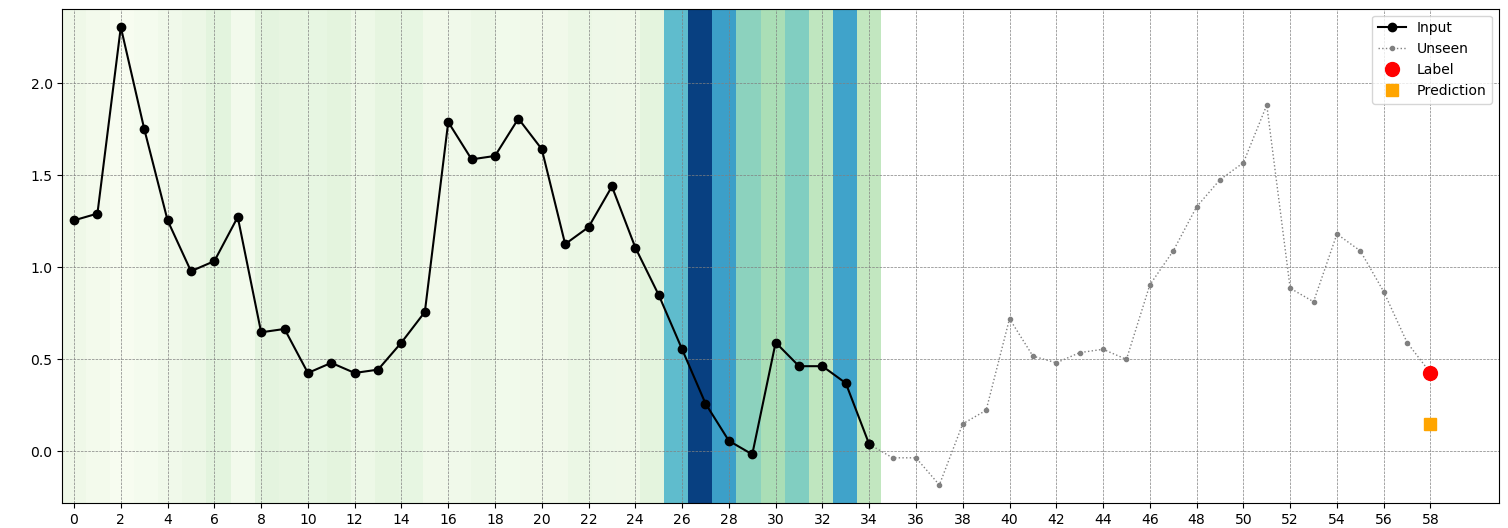}
  \caption{Informer trained with MSE 
  	}\label{fig:subc}
\end{subfigure} 
\begin{subfigure}[]{.49\textwidth}
  \centering
  \includegraphics[width=\linewidth]{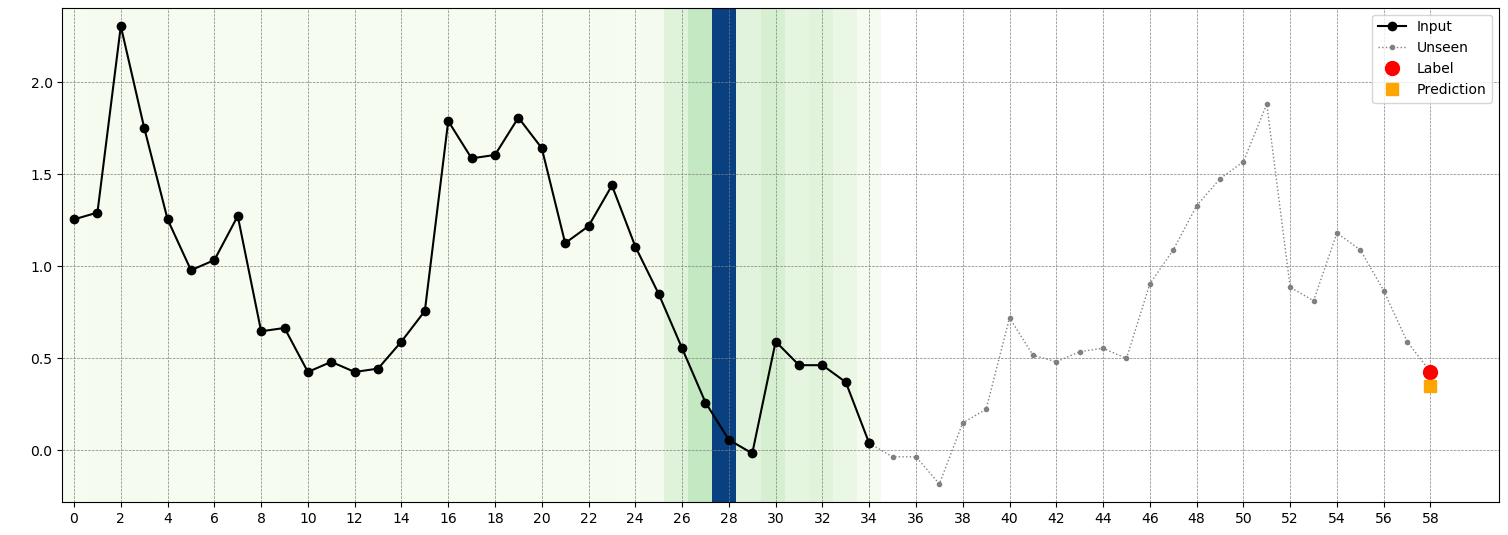}
  \caption{Informer trained with MSE and MoCo2 
  	}\label{fig:subd}
\end{subfigure} 

\begin{subfigure}[]{.48\textwidth}
	\centering
	\includegraphics[width=\linewidth]{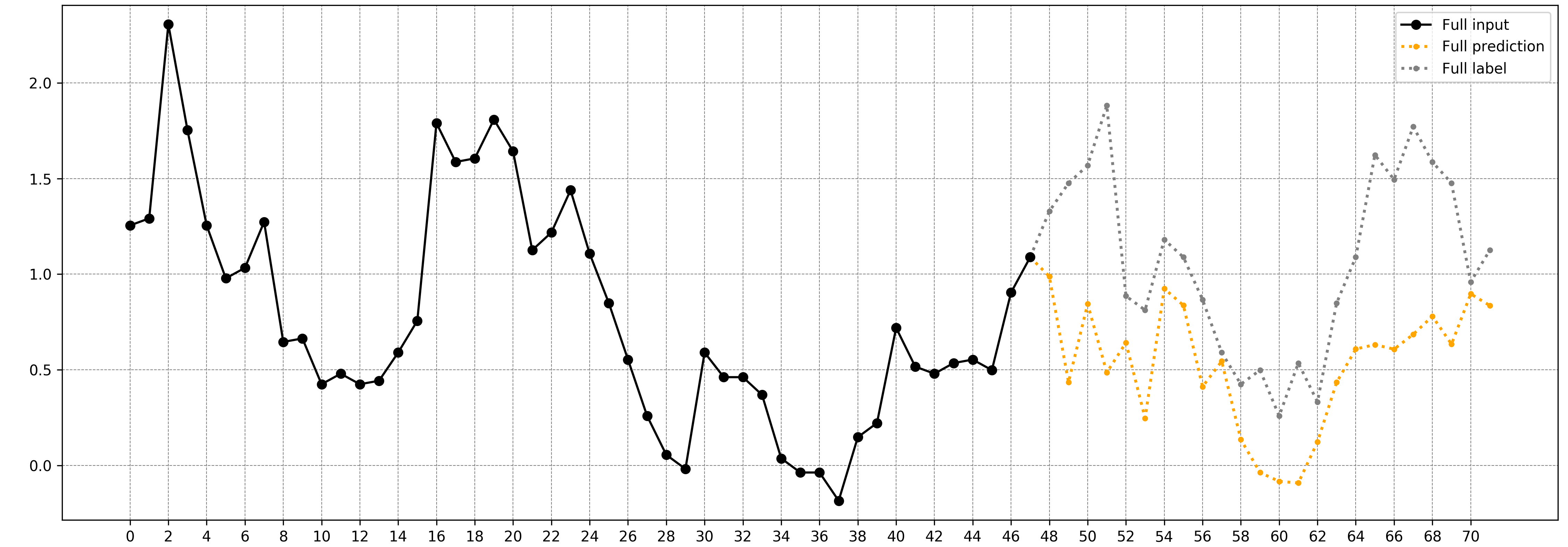}
	\caption{TCN w. MSE and MoCo2 long-term prediction.
	}\label{fig:sube}
\end{subfigure} 
\begin{subfigure}[]{.48\textwidth}
	\centering
	\includegraphics[width=\linewidth]{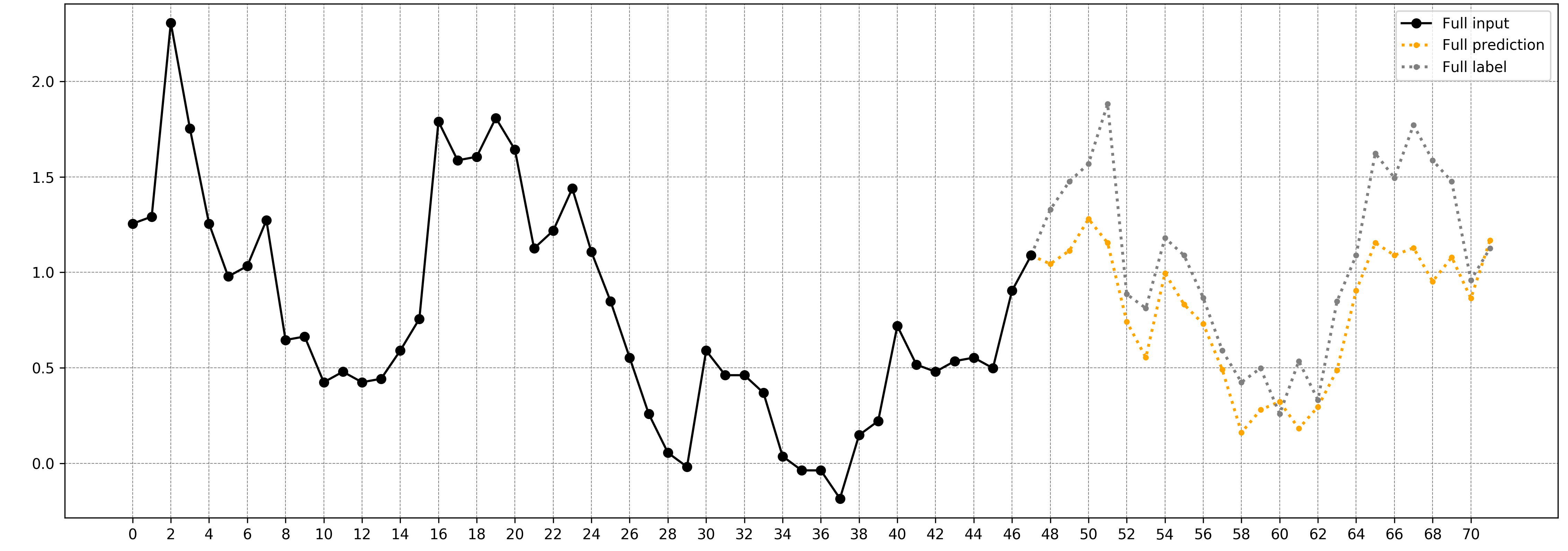}
	\caption{Informer w. MSE and MoCo2 long-term prediction.
	}\label{fig:subf}
\end{subfigure} 
\vspace{-0.1in}
\caption{Comparison of effective receptive field and long-term prediction of TCNs and Informers on ETTh1 dataset with a prediction length of $24$.
Denser color indicates larger gradient to the input timestamp. 
}
\label{fig:tcn_rf}
\vspace{-0.2in}
\end{figure}

\subsection{Qualitative Analysis}
To gain deeper insights into how SSCL enhances time-series forecasting, we conduct a qualitative analysis of the empirical receptive field (RF) of a TCN, following the approach proposed by~\cite{luo-2016-understanding}.
To carry out this analysis, we pass a validation set sample $x_i$ through the trained model and calculate the MSE loss between the model prediction $\hat{y}_i^j$ and the corresponding gold label $y_i^j$ at timestamp $j$, using Eq.~\ref{eq:mse}. We then calculate the gradient of the input sequence $x_i$ with respect to the loss by applying back-propagation, which is formulated as:

\begin{equation}
\frac{\partial \mathcal{L}_i^j}{\partial x_i} = \frac{\partial(y_i^j-\hat{y}_i^j)^2}{\partial x_i}
\label{eq:gradient}
\end{equation}

As illustrated in Figure~\ref{fig:tcn_rf}, SSCL allows both the TCN and Transformer models to concentrate more on related information, such as scale or periodic relationships, for forecasting, as compared to the vanilla models trained exclusively with MSE loss.
Our analysis of Figure~\ref{fig:subb} and Figure~\ref{fig:subd} demonstrates that the models trained with SSCL exhibit a significant focus on the timestamp that exhibits a similar periodic pattern as the target timestamp. 
Furthermore, the TCN and Transformer models, trained with SSCL, effectively capture historical information associated with the scale or periodic pattern of the target timestamp. 
These findings underscore the effectiveness of SSCL in improving the performance of time-series forecasting models.
Note that we only visualize the effective receptive field in Eq.~\ref{eq:gradient} for a single time step prediction to avoid cluttered plots.
We also draw a comparison of long-term predictions as depicted in Figure~\ref{fig:sube} and Figure~\ref{fig:subf}.

\section{Conclusion}
In this study, we scrutinize the impact of various training variables on the learning process for time series forecasting with SSCL.
Our exhaustive investigation encompasses various SSCL algorithms, learning strategies, model architectures, and their interplay.
Through our analysis, we derive the following key findings: (1) For leveraging SSCL in learning time series representations, Transformer based model demonstrates better performance than TCN and LSTM based models. (2) Among the SSCL algorithms evaluated, MoCo2 emerges as the most effective in terms of performance. (3) Incorporating SSCL as an auxiliary objective within an end-to-end learning framework yields the best results for time series forecasting tasks. (4) Importantly, our experiments reveal that a Transformer model, trained end-to-end with MSE loss in tandem with SSCL, achieves the most notable improvement.
Furthermore, our qualitative analyses shed light on the capabilities of SSCL in capturing historical information pertaining to the scale or periodic patterns of the target timestamp. 
We hope that these insights offer valuable guidance for utilizing SSCL in time series forecasting and also ignite and facilitate future research endeavors in this domain.

\clearpage
\bibliography{literature}
\bibliographystyle{acl_natbib}

\appendix

\end{document}